\documentclass[letterpaper, 10 pt, conference]{ieeeconf}  

\usepackage{pifont}   
\usepackage{amsmath}
\usepackage{mathtools,amsfonts} 

\usepackage{xcolor}
\usepackage{array}                        
\usepackage{multirow}                     
\usepackage{graphicx}                     
\usepackage{color, colortbl}              
\usepackage{booktabs}                     
\usepackage{xcolor}
\usepackage{gensymb} 					 
\usepackage{hyperref}
\usepackage{cite}

\usepackage{algorithm}
\usepackage{algorithmic}

\IEEEoverridecommandlockouts                              
\title{\LARGE \bf
	Enhancing Hand Palm Motion Gesture Recognition by Eliminating Reference Frame Bias via Frame-Invariant Similarity Measures
}

\author{Arno Verduyn, Maxim Vochten, and Joris De Schutter
\thanks{All authors are with the Department of Mechanical Engineering and Flanders Make at KU Leuven, 3001 Leuven, Belgium.}%
}

\begin{document}

\maketitle
\thispagestyle{empty} 
\pagestyle{empty} 

\begin{abstract}

The ability of robots to recognize human gestures facilitates a natural and accessible human-robot collaboration. However, most work in gesture recognition remains rooted in reference frame-dependent representations. This poses a challenge when reference frames vary due to different work cell layouts, imprecise frame calibrations, or other environmental changes. This paper investigated the use of invariant trajectory descriptors for robust hand palm motion gesture recognition under reference frame changes. 
First, a novel dataset of recorded Hand Palm Motion (HPM) gestures is introduced. The motion gestures in this dataset were specifically designed to be distinguishable without dependence on specific reference frames or directional cues. 
Afterwards, multiple invariant trajectory descriptor approaches were benchmarked to assess how their performances generalize to this novel HPM dataset. 
After this offline benchmarking, the best scoring approach is validated for online recognition by developing a real-time Proof of Concept (PoC). In this PoC, hand palm motion gestures were used to control the real-time movement of a manipulator arm. The PoC demonstrated a high recognition reliability in real-time operation, achieving an $F_1$-score of 92.3\%.
This work demonstrates the effectiveness of the invariant descriptor approach as a standalone solution. Moreover, we believe that the invariant descriptor approach can also be utilized within other state-of-the-art pattern recognition and learning systems to improve their robustness against reference frame variations.
\end{abstract}


\section{Introduction}
\label{sec:intro}

The development and acceptance of collaborative robots in industry is highly dependent on natural, accessible, and reliable human-robot interaction (HRI) \cite{neto2019gesture}. A key step toward natural HRI is enabling robots to recognize both spoken words and human gestures. This paper focuses on the recognition of dynamic gestures (i.e. hand palm motions).

\subsection{Problems in Gesture Recognition}
Most current work in human gesture (or action) recognition remains rooted in view-dependent representations~\cite{WEINLAND}.  Such representations do not extrapolate well toward other viewpoints, limiting recognition in changing situations. This issue has become a bottleneck for human motion analysis and its applications~\cite{IEEE2010}. In addition, the lack of multi-viewpoint datasets also sets up barriers~\cite{IEEEArbitraryView}. The works in \cite{WEINLAND,IEEE2010,IEEEArbitraryView,rahmani2016,IEEEVideo,Holte2011multiview,IEEE2023} aim to address these issues. However, most approaches solely focus on solving the \textit{camera viewpoint} issue in \textit{vision-based} human motion recognition.

Besides vision-based motion recognition, it is also common to use marker- or tracker-based 3D motion measurement systems because of their higher recording accuracy. For those systems, in addition to variation in the sensor viewpoint, there can also be variation in the \textit{body reference frame} on the moving body. This reference frame, rigidly attached to the moving body, defines the body's relative position and orientation with respect to a sensor frame. If the sensor frame remains stationary relative to the environment, the sensor frame is commonly referred to as a \textit{world reference frame}. The world and body reference frames can significantly vary across motions due to different work cell layouts in flexible manufacturing, imprecise frame calibrations during setup initialization, inconsistent marker or tracker configurations, or other environmental factors. 

Other variations that occur when performing motions in different circumstances include changes in the execution speed (or `time profile') of the motion, partial occlusion of the motion data due to marker occlusion, and incomplete motion data in real-time scenarios. 

\begin{figure}
	\centering
	\medskip
	\includegraphics[width=0.75\linewidth]{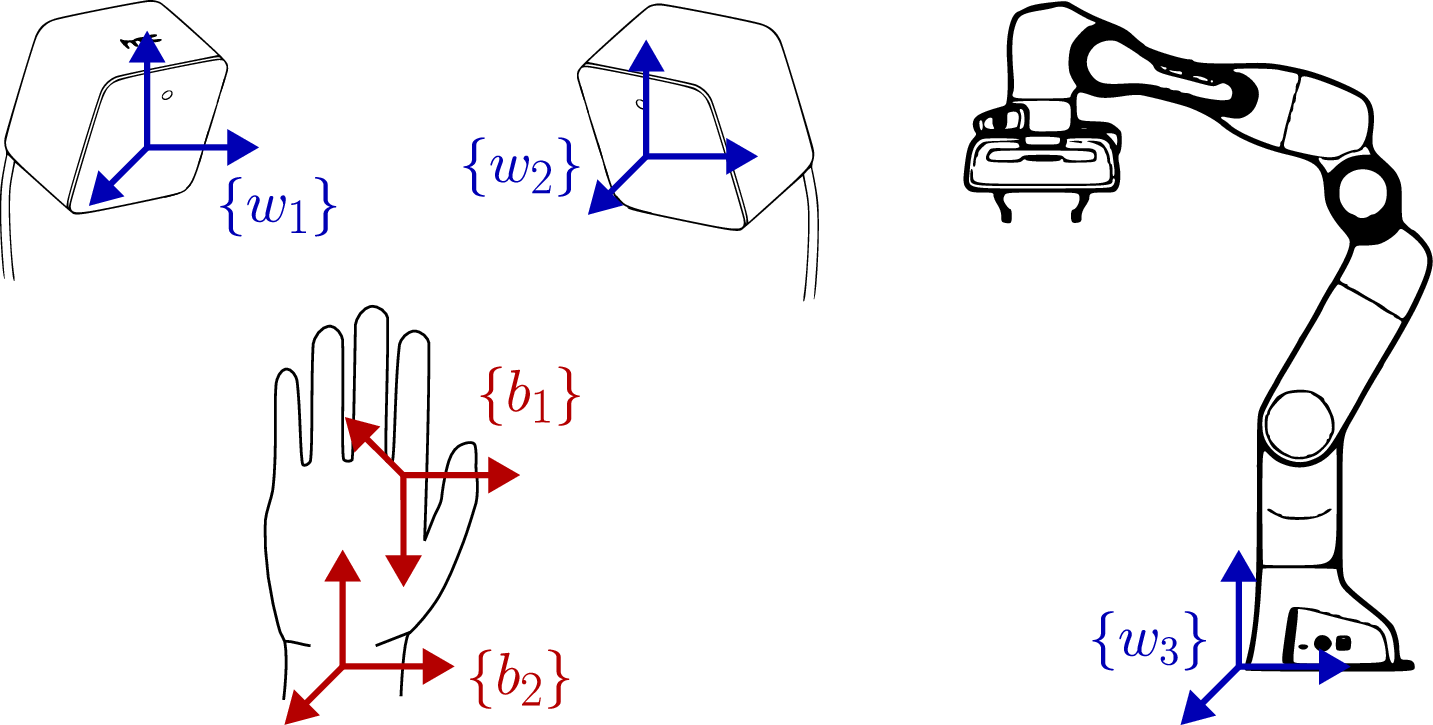}
	\caption{Visualization of different reference frames when describing hand palm motion gestures in human-robot interaction using a motion capture setup. The world reference frame often depends on the setup of the measurement system. Here, two HTC Vive base stations are shown, with two possible world frames, $\{w_1\}$ and $\{w_2\}$, depicted in blue. Another choice for the world frame is the base frame of the robot, $\{w_3\}$. The body reference frame often depends on the pose of the tracker attached to the moving body. Here, two possible body frames, $\{b_1\}$ and $\{b_2\}$, describing the relative pose of the hand with respect to a world reference frame, are depicted in red. Different choices for these frames will result in different motion trajectory coordinates of the performed gestures. This limits the effective  recognition of the gestures when these frames differ between setups or change in real-time.
	\textit{The line art of the Franka Panda robot and the HTC Vive tracker and base stations were retrieved from their respective online manuals.}}
	\label{fig:frame_variations}
	\vspace{-5pt}
\end{figure}

In this paper, variations in the world frame, body frame, time profile, as well as differences in data occlusion, are referred to as \textit{contextual variations}. When such contextual variations occur, the effective recognition of motions becomes challenging. This challenge arises since the trajectory coordinates of a motion depend on the context and will therefore change when the context varies.

This paper addresses the specific problem of enabling robots to recognize hand palm motion gestures when challenged by contextual variations. These hand palm motion gestures are represented by the 3D spatio-temporal trajectories of the palm of the hand, hence excluding finger motions. The gestures involve both rotation and translation of the hand. The motion of the hand is captured using an HTC Vive motion capture system. In this setup, different choices for the world and body reference frames exist, and different choices for these frames will result in different motion trajectory coordinates. Various choices for these reference frames are visualized in Figure~\ref{fig:frame_variations}. 

\subsection{Approaches for dealing with contextual variations}
To deal with contextual variations when recognizing motions, approaches involving \textit{context alignment}, \textit{machine learning}, and \textit{invariant descriptors} have been proposed.

\textit{Context alignment approaches} aim to remove contextual variations using alignment strategies. Frames can be spatially aligned using affine transformations \cite{wang2008manifold,kabsch1976solution,Rao2002,DSRF2018}. Time profiles can be aligned using algorithms like Dynamic Time Warping (DTW) \cite{sakoe1978dynamic}. However, aligning frames and time profiles simultaneously remains computationally expensive~\cite{FlexibleSyntacticMatching1999}, limiting this simultaneous approach to offline applications.

\textit{Machine learning approaches}~\cite{MLreview,ViewInvariantANN2012} aim to identify salient patterns in trajectories that recur across diverse contexts. However, a major challenge lies in preventing these approaches from erroneously identifying patterns within irrelevant contextual features, which can happen when the dataset for training is sparse or biased toward specific contexts~\cite{BiasMitigation}. A potential solution to this challenge involves constructing a dataset that represents a wide range of contexts in an unbiased manner~\cite{rahmani2016,Holte2011multiview}, but this approach can be labor-intensive and may still fall short of capturing all possible contextual variations. 

\textit{Invariant descriptor approaches}~\cite{Lee2018, DeSchutterJoris2010,Vochten2015,vochten2023invariant} aim to remove the context from the trajectory data to obtain context-invariant trajectory descriptors. Invariant descriptor approaches have shown superior extrapolation capabilities toward unrepresented context, even when the dataset at hand is sparse\cite{Vochten2015}. Section~\ref{sec:related_work} provides a brief overview of invariant descriptor approaches.

\subsection{Paper Objectives and Contributions}

The contributions of this paper are threefold. 
First, to aid researchers and industrial practitioners in developing recognition systems that are robust against contextual variations, a novel dataset of hand palm motion gestures is introduced. The gestures were specifically designed to be distinguishable without dependence on specific reference frames or directional cues. To challenge gesture recognition, the dataset was augmented with artificial variations, including variations in reference frames, execution speeds, and data occlusion.

Second, offline benchmarking is used to evaluate how the findings in \cite{verduyn2025} generalize to this novel HPM dataset, while also identifying the most suitable invariant descriptor approach for this dataset, which is found to be BILTS$^+$.

Third, the BILTS$^+$ approach is validated for online recognition by developing a real-time Proof of Concept (PoC). In this PoC, hand palm motion gestures were used to control the real-time movement of a manipulator arm. The PoC demonstrated a high recognition reliability in real-time operation, achieving an $F_1$-score of 92.3\% despite reference frame variations. That is, the user is free to move around and hold the HTC Vive tracker in arbitrary ways without compromising recognition accuracy. Furthermore, the approach's frame-invariance eliminated the need for calibration between the tracker frame, sensor frame, and robot base frame, which significantly simplified setup deployment.

\section{Related Work on Invariant Descriptors}
\label{sec:related_work}

The foundational Frenet-Serret (FS)~\cite{Frenet1852,Serret1853} formulas describe \textit{point curves} in an invariant manner by describing their \textit{local shape} instead of their trajectory coordinates. This local shape is defined in terms of differental-geometric properties (curvature and torsion) expressed in a local moving frame (the FS frame). The extended FS (eFS)~\cite{Vochten2015} and Denavit-Hartenberg Bidirectional (DHB)\cite{Lee2018} descriptors extend this approach toward \textit{rigid-body motion}. This extension consists of the definition of two FS frames, one for the rotation and one for the translation of the moving body. 

The eFS and DHB descriptors are \textit{left-invariant}\cite{Park1995}, meaning they are invariant to changes in the world frame. However, they are not \textit{right-invariant}\cite{Park1995}, meaning they are not invariant to changes in the body frame. That is, they require the user to choose and calibrate a body reference point to define the body's translational motion. Another important drawback of the eFS and DHB descriptors is the loss of the internal kinematic relations between the rotational and translational motion of the moving body due to the introduction of two decoupled FS frames.

The above explained issues can be eliminated by making use of \textit{screw theory}~\cite{davidson2004robots} such that the descriptor becomes both left- and right-invariant, also referred to as \textit{bi-invariant}~\cite{Park1995}. From the work of Chasles \cite{chasles1830note}, it is known that the first-order kinematics of a moving body can be described in a bi-invariant way by its rotation and translation along the \textit{Instantaneous Screw Axis (ISA)} of the motion. Based on this concept, De Schutter \cite{DeSchutterJoris2010} introduced the ISA descriptor, consisting of six numbers, which describes the first-order kinematics of both the moving body and the moving ISA in a bi-invariant manner. 

\begin{figure*}[t!]
	\centering
	\medskip
	\includegraphics[width=0.85\linewidth]{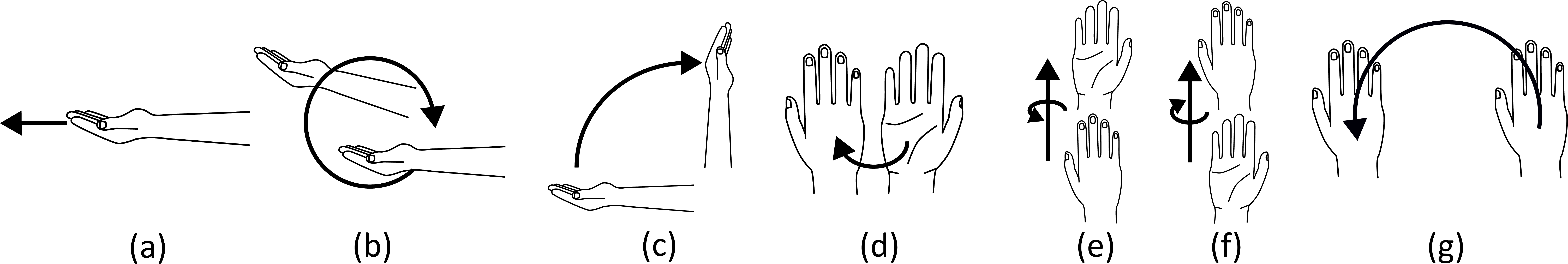}
	\caption{Visualization of the hand palm motions within the HPM dataset: (a) \textit{Go Left}: move the hand along a straight line, (b) \textit{Go Right}: trace a circular path with the hand while allowing a slight rotation of the forearm, (c) \textit{Go Up}: move the hand upward while rotating the forearm, (d) \textit{Go Down}: rotate the hand so the palm transitions from facing upward to facing downward, (e) \textit{Open Gripper}: starting from a downward facing palm, move the hand forward while rotating it such that the palm faces upward (this generates a screw motion with a positive pitch), (f) \textit{Close Gripper}: starting from an upward facing palm, move the hand forward while rotating it such that the palm faces downward (this generates a screw motion with a negative pitch). (g) \textit{Go Home}: trace an arc by translating the hand.}
	\label{fig:gestures}
	\vspace{-5pt}
\end{figure*}

\textit{Global} descriptor approaches, such as the Rotation and Relative Velocity (RRV) descriptor approach~\cite{RRV2018} and the Dual Square-Root Function (DSRF) descriptor approach~\cite{DSRF2018} aim to obtain a higher noise robustness by introducing global trajectory features or global trajectory alignment strategies, thereby sacrificing true locality. However, the sacrifice of true locality can result in higher computational overhead within online applications when combined with alignment algorithms like DTW over a sliding window~\cite{verduyn2025}. Furthermore, both the RRV and DSRF approaches lack right-invariance. 

A common challenge among the above explained approaches is the severe noise sensitivity near \textit{singularities}, where the invariant descriptors are not uniquely defined. To address this issue, the Bi-Invariant Local Trajectory-shape Similarity (BILTS) approach has been proposed more recently~\cite{verduyn2025}. This approach is bi-invariant and showed exceptional robustness to noise near singularities. The BILTS approach describes the kinematics of the moving body in a bi-invariant manner using a local moving frame defined on the ISA, similar to the ISA descriptor, but now using a non-minimal descriptor of 14 numbers, enhancing its robustness to noise near singularities. In~\cite{verduyn2025}, additional regularization actions were introduced (BILTS$^+$) to further strengthen the approach's robustness to noise near singularities.

Section~\ref{sec:comp} compares all the above explained invariant descriptor approaches for the recognition of the gestures within the HPM dataset introduced in the next section.

\section{Hand Palm Motion (HPM) dataset}

This section introduces the novel Hand Palm Motions (HPM) dataset\footnote{Dataset available on Zenodo: \\ \url{https://doi.org/10.5281/zenodo.15020057}}. The motions within the HPM dataset involve rotational movements, translational movements, or a combination of both performed by the palm of the hand. These motions were designed for the PoC detailed in Section~\ref{sec:PoC}. In this PoC, the goal is to control the movement of the end-effector and gripper fingers of the Franka Panda robot through hand palm motion gestures. 
Figure~\ref{fig:gestures} visualizes the seven gestures that were designed: \textit{Go Left, Go Right, Go Up, Go Down, Open Gripper, Close Gripper,} and \textit{Go Home}. These motion gestures are easy to perform, which ensures accessibility for users. Additionally, the gestures were specifically designed such that distinguishing the gestures does not rely on specific coordinate reference frames or directions. That is, the \textit{shape} of the motion (rectilinear or circular translation, pure rotation, pure screw motion, etc.) contains sufficient information for this distinction.

To reduce transition effects between gestures that are performed successively, all gestures, except for \textit{Go Right}, were designed to include the motions as shown in Figure~\ref{fig:gestures}, followed by their \textit{reverse} motions. Hence, after each gesture, the hand returned to the same pose it started from.

The HPM dataset was designed to consist of simple hand motions to ensure ease of execution for users. However, these simple motions do not excite all degrees of freedom within 3D free space motion. Such motions often contain local \textit{singularities}, where the invariant descriptors are not uniquely defined. Hence, by design, the dataset inherently presents a significant challenge for invariant descriptor approaches, allowing to test their robustness against singularities.

The hand palm motions were recorded using an HTC Vive motion capture system, where the user's hand motion was captured by holding an HTC Vive tracker. The HTC Vive system recorded the orientation and position of the tracker with an accuracy of a few degrees and a few millimeters, respectively. The orientation and position trajectories of the tracker were retained as quaternion coordinates and 3D position coordinates sampled at a frequency of 50 Hz. For each of the seven gestures, five trials were recorded, resulting in a total of $7\times5=35$ recordings.

To introduce the challenge of dealing with contextual variations when recognizing motions, the HPM dataset is augmented toward 420 trials by artificially transforming and perturbing the recordings. 

Twelve different contexts were designed.
The context \textit{Original 1} consists of the original recordings. The context \textit{Original 2} serves as a baseline, with no artificial transformations applied. The other contexts consist of artificially transformed versions of the trials from \textit{Original 1}. To prevent that the trials from the context \textit{Original~2} `exactly' match those from the context \textit{Original~1}, small perturbations were introduced by adding white noise with standard deviations of 1 mm and 1$\degree$ to the position and orientation trajectories, respectively. For consistency reasons, this noise perturbation was applied to every trial of each context.
Consequently, trajectory preprocessing techniques and recognition algorithms that can handle noise effectively are expected to perform well when recognizing the motions in the HPM dataset. 

The contexts \textit{Slower} and \textit{Faster} were obtained by rescaling the time axis and numerically resampling the trajectory coordinates using Screw Linear Interpolation (ScLERP)~\cite{kavan2008geometric}, a generalization of SLERP to $SE(3)$. The resulting transformed trajectories simulate twice as slow and twice as fast executions of the gestures. 

The context \textit{First Half} includes trajectories that consist of only the first half of the trajectory data. These trajectories hence represent gestures that have not yet been finished. This context allows the evaluation of an approach's ability to recognize trajectories when dealing with incomplete data. 

The six contexts \textit{Change in body frame 1-3} and \textit{Change in world frame 1-3} incorporate reference frame transformations. To apply these transformations, the trajectory representation was first transformed from the quaternion and position representation $(\boldsymbol{q}(t_k),\boldsymbol{p}(t_k))$ toward the homogeneous matrix representation $\boldsymbol{T}(t_k)$, with $t_k$ the time at sample $k$.  Transformations of the body and world frames were then achieved by right- or left-multiplication~\cite{murray1994mathematical} of the trajectories $\boldsymbol{T}(t_k)$ with a constant homogeneous transformation matrix, respectively. The resulting transformations are the following: 

\begin{itemize}
	\item \textit{Change in body frame 1}: $\{b\}$ is translated along its \mbox{$x$-axis} with 5 cm and rotated about its $z$-axis with 180\degree.
	\item \textit{Change in body frame 2}: $\{b\}$ is translated along its \mbox{$y$-axis} with 5 cm and rotated about its $x$-axis with 90\degree.
	\item \textit{Change in body frame 3}: $\{b\}$ is translated along its \mbox{$z$-axis} with \hspace{-1pt}-5 \hspace{-1pt}cm and rotated about its $y$-axis with \hspace{-2pt}-90\degree.
	\item \textit{Change in world frame 1}: $\{w\}$ is translated along its \mbox{$x$-axis} with 1 m and rotated about its $z$-axis with 180\degree.
	\item \textit{Change in world frame 2}: $\{w\}$ is translated along its \mbox{$y$-axis} with 1 m and rotated about its $x$-axis with 90\degree.
	\item \textit{Change in world frame 3}: $\{w\}$ is translated along its \mbox{$z$-axis} with 1 m and rotated about its $y$-axis with -90\degree.
\end{itemize}

The origin of the body frame was translated by only 5 cm. Hence, this perturbation remained within reasonable deviations with respect to the size of the human hand.  

The context \textit{Combination} incorporates multiple transformations. That is, the motions were simulated to be performed twice as fast, only the first half of the trajectory data was retained, and both the body and world frames were varied.

\section{Experimental Comparative Analysis}
\label{sec:comp}
To robustly recognize the motions within the HPM dataset, which is rich in contextual variations, invariant descriptor approaches can be leveraged. 
In \cite{verduyn2025}, different invariant descriptor approaches were implemented and their perfomances were compared for the recognition of object manipulation tasks. The goals of the comparative experiments in this section are to reconfirm the findings in \cite{verduyn2025}, assess how those findings generalize to this novel HPM dataset, and identify the most suitable invariant descriptor approach for this HPM dataset. To achieve these goals, the same experimental procedure as in \cite{verduyn2025} was employed, which is briefly described below. For more detailed explanations, please refer to \cite{verduyn2025}. 

\subsection{Experimental design and results}

\textit{Calculation of invariant descriptors}: The time dependency of the trajectories was first removed by transforming the temporal trajectories to geometric trajectories using numerical reparameterization. This allows trajectories to be classified solely based on spatial geometry. Different choices for the geometric progress parameter were compared, including the traversed \textit{arclength} traced by the body reference point~\cite{DSRF2018,RRV2018,Lee2018}, the traversed \textit{angle} of the moving body~\cite{Roth2005}, and the traversed \textit{screw path} of the moving body~\cite{verduyn2023enhancingmotiontrajectorysegmentation}. This screw path can be interpreted as the time integral of the combined rotational and translational velocity of the moving body along the ISA of the motion.  From the reparameterized trajectories, invariant descriptors were calculated using the analytical formulas provided in~\cite{DSRF2018,RRV2018,Vochten2015,DeSchutterJoris2010,Lee2018,verduyn2025}. 

\textit{Parameter tuning and motion classification}:
A dedicated \textit{training} and \textit{testing} routine was followed to train the internal parameters of the different invariant descriptor approaches and afterwards evaluate their recognition performance when applied to unseen data. 
The invariant descriptors of the trials from the context \textit{Original 1} served as the neighbors within a 1-Nearest-Neighbor (\mbox{1-NN}) classifier. The first two trials of each context apart from the context \textit{Original 1} served as the training set. Using this training set, values for the parameters of each approach were determined using grid search over a predefined grid. The highest Recognition Ratio (RR) during training served as an objective criterion for parameter training. After parameter training, the remaining trials within the dataset were classified using the same \mbox{1-NN} classifier. To deal with small misalignments in geometric progress of the descriptors during classification, DTW alignment of the descriptors was applied.
The test results are reported in Table~\ref{tab:test_results_HPM}, including the RR per context, the total average RR, and the standard deviation $\sigma$ in RR across the different contexts. Table~\ref{tab:test_results_HPM} also introduces the notation $\Delta \{b\}$ and $\Delta\{w\}$ to concisely denote the \textit{Change in body frame 1-3} and \textit{Change in world frame 1-3} contexts, respectively.

\definecolor{mylightblue}{RGB}{235, 235, 255}
\definecolor{mydarkblue}{RGB}{170, 170, 255}

\begin{table}[t]
	\centering
	\caption{Recognition results (\%) for the Hand Palm Motions dataset.}
	\label{tab:test_results_HPM}	
	\resizebox{\linewidth}{!}{%
		\renewcommand{\arraystretch}{1}
		\begin{tabular}{llccccccc>{\columncolor{mylightblue}}l}
			\toprule 
			\hspace{-5pt}\textbf{method} & \textbf{progress} &
			\hspace{-7pt}\textbf{~~original\hspace{1pt}2~}\hspace{-5pt} & \hspace{-4pt}\textbf{slower}\hspace{-5pt} &  \hspace{-5pt}\textbf{faster}\hspace{-8pt} & \hspace{-5pt}\textbf{half}\hspace{-5pt} & \hspace{-10pt}$\boldsymbol{\Delta \{b\}}$\hspace{-10pt}  & \hspace{-10pt}$\boldsymbol{\Delta \{w\}}$\hspace{-10pt} & \hspace{-5pt}\textbf{comb.}\hspace{-5pt} &\textbf{mean$\pm\sigma$}\\
			\addlinespace
			& arclength & 85.7 & 85.7 & 100 & 57.1 & 84.1 & 85.7 & 47.6 & 80.5$\pm14.8$\hspace{-5pt} \\
			\hspace{-7pt}DHB~\cite{Lee2018} & angle & 66.7 & 66.7 & 95.2 & 42.9 & 60.3 & 63.5 & 57.1 & 63.6$\pm13.0$\hspace{-5pt} \\
			& screw path\hspace{-10pt} & 85.7 & 71.4 & 100 & 61.9 & 88.9 & 84.1 & 71.4 & 82.7$\pm10.7$\hspace{-5pt} \\
			\addlinespace
			& arclength & 85.7 & 71.4 & 95.2 & 42.9 & 74.6 & 85.7 & 47.6 & 74.9$\pm16.5$\hspace{-5pt} \\
			\hspace{-7pt}eFS~\cite{Vochten2015} & angle & 71.4 & 76.2 & 95.2 & 66.7 & 66.7 & 65.1 & 71.4 & 70.6$\pm9.2$\hspace{-5pt} \\
			& screw path\hspace{-10pt}  & 100 & 90.5 & 100 & 66.7 & 95.2 & 98.4 & 38.1 & 88.7$\pm19.3$\hspace{-5pt} \\
			\addlinespace
			\hspace{-7pt}\multirow{2}{*}{ISA~\cite{DeSchutterJoris2010}} & angle & 71.4 & 61.9 & 85.7 & 57.1 & 71.4 & 74.6 & 61.9 & 70.5$\pm7.9$\hspace{-5pt} \\
			& screw path\hspace{-10pt} & 85.7 & 81.0 & 100 & 95.2 & 85.7 & 85.7 & 81.0 & 87.0$\pm5.7$\hspace{-5pt} \\
			\addlinespace
			\hspace{-7pt}\multirow{2}{*}{BILTS~\cite{verduyn2025}}\hspace{-10pt}& angle & 76.2 & 52.4 & 90.5 & 66.7 & 74.6 & 74.6 & 90.5 & 74.9$\pm10.7$\hspace{-5pt} \\
			& screw path\hspace{-10pt} & 95.2 & 85.7 & 100 & 100 & 98.4 & 98.4 & 100 & 97.4$\pm4.5$\hspace{-5pt} \\
			\addlinespace
			\hspace{-7pt}\textbf{BILTS$^+$}\hspace{-0.1pt}\cite{verduyn2025}\hspace{-10pt} & screw path\hspace{-10pt} & \textbf{100} & \textbf{100} & \textbf{100} & \textbf{100} & \textbf{100} & \textbf{100} & \textbf{100} & \textbf{100}$\boldsymbol{\pm0.0}$\hspace{-5pt} \\
			\addlinespace
			& arclength & 81.0 & 90.5 & 95.2 & 61.9 & 82.3 & 85.4 & 42.9 & 79.5$\pm14.9$\hspace{-5pt} \\
			\hspace{-7pt}RRV~\cite{RRV2018} & angle  & 66.7 & 61.9 & 61.9 & 57.1 & 69.8 & 66.7 & 80.9 & 67.1$\pm8.4$\hspace{-5pt} \\
			& screw path\hspace{-5pt} & 66.7 & 47.6 & 71.4 & 33.3 & 61.9 & 69.9 & 33.3 & 58.9$\pm15.4$\hspace{-5pt} \\
			\addlinespace
			& arclength & 100 & 100 & 100 & 57.1 & 100 & 100 & 28.6 & 89.6$\pm24.0$\hspace{-5pt} \\
			\hspace{-7pt}DSRF~\cite{DSRF2018} & angle & 100 & 90.5 & 100 & 71.4 & 100 & 100 & 47.6 & 91.8$\pm17.1$\hspace{-5pt} \\
			& screw path\hspace{-5pt} & 100 & 100 & 100 & 38.1 & 96.8 & 100 & 38.1 & 87.9$\pm24.8$\hspace{-5pt} \\
			\addlinespace
			\bottomrule
		\end{tabular}
	} \vspace{-5pt}
\end{table}

\subsection{Consistency with previous results}

Compared to the results reported in \cite{verduyn2025}, the new results in Table~\ref{tab:test_results_HPM} reconfirm the superiority of BILTS$^+$, the robustness of bi-invariant approaches, and the superiority of the screw-based-progress when dealing with contextual variations.

For the HPM dataset, the BILTS$^+$ approach outperformed the other invariant descriptor approaches.  Even for the context \textit{Combined}, where the gestures were performed twice as fast, both world and body frames were varied, and only 50\% of the trajectory data was available, the BILTS$^+$ approach still achieved a perfect score. 
The BILTS$^+$ approach obtained this superior performance since it simultaneously provided:
\begin{enumerate}
	\item a \textit{bi-invariant} similarity measure, providing a high robustness to both world and body frame variations.
	\item a \textit{local} similarity measure, providing a higher robustness in the case of incomplete data,
	\item a \textit{complete} similarity measure, since it is based on a \textit{complete} third-order trajectory-shape description~\cite{verduyn2025}, which allows for more effective DTW matching,
	\item a \textit{high robustness against singularities}, allowing to robustly deal with `simple' motions that do not excite all degrees of freedom within a free space motion. 
\end{enumerate}

Motivated by the superior performance of BILTS$^+$ for the HPM dataset, the BILTS$^+$ approach is used for the online gesture recognition PoC, which is detailed in Section~\ref{sec:PoC}.

The local bi-invariant approaches (ISA, BILTS and BILTS$^+$) demonstrated high robustness to contextual variations, reflected by their relatively low standard deviations of $5.7\%$, $4.5\%$, and $0.0\%$, respectively.

Lastly, among local approaches (DHB, eFS, ISA, BILTS, BILTS$^+$), \textit{screw path} once again proved to be the best geometric progress type, consistently yielding higher average recognition rates (RRs) compared to \textit{arclength} or \textit{angle}.

\subsection{Differences from previous results}
Compared to the results reported in \cite{verduyn2025}, notable differences with the results for the HPM dataset can be identified:

\vspace{3pt}
\textit{1) DSRF no longer outperformed BILTS}:
In previous results, the DSRF approach outperformed the unregularized BILTS approach. This is not the case for the HPM dataset, since BILTS (97.4\%) outperformed DSRF (91.8\%). 

The average RR for the DSRF approach was significantly affected by the relatively low RRs for the contexts \textit{First Half} and \textit{Combined}. These contexts allowed the evaluation of an approach's ability to recognize incomplete trajectories. This challenge was not present in the datasets in \cite{verduyn2025}. Global methods (such as DRSF and RRV) are expected to underperform when dealing with this challenge, as they rely on global trajectory features, which are impossible to determine accurately when incomplete trajectory data is provided. 

Compared to DSRF, BILTS performed especially well when recognizing the motions from the contexts \textit{First Half} and \textit{Combined}. To achieve high RRs for these contexts, effective DTW matching of the incomplete trials with the first parts of the original trials was essential. The BILTS approach achieved RRs of 100\% for these contexts. These high RRs can be attributed to the BILTS descriptor's \textit{complete} third-order trajectory-shape description~\cite{verduyn2025}, a unique advantage compared to DHB, eFS, and ISA. This complete shape information likely led to effective local DTW matching, and consequently, also led to higher RRs.

\vspace{3pt}
\textit{2) Angle-based progress consistently underperformed}: 
Another new finding is that, for the local approaches, using \textit{angle} as the geometric progress consistently underperformed compared to \textit{arclength} and \textit{screw path}. Using the traversed angle of the hand as the geometric progress was likely not a good choice for the HPM dataset since some of the gestures involved translational motions of the hand. For these motions, the traversed angle of the hand does not represent a meaningful progress, since it primarily represents noise and small irrelevant human variations.

\vspace{3pt}
\textit{3) Lower sensitivities to body frame variations}:
Unlike the results in \cite{verduyn2025}, approaches that lack right-invariance (DHB, eFS, RRV, DSRF) were not drastically affected by the changes in the body frame. In \cite{verduyn2025}, the origin of the body frame was artificially varied by magnitudes up to 30cm, a magnitude that is reasonable when describing the motion of larger objects like bowls, mixing containers, pans with handles, water jugs, brooms, etc. In the HPM dataset, the body frame displacements were much smaller (5cm), corresponding to the size of the human hand. These small displacements hence did not cause severe detriments in the RRs for the approaches lacking right-invariance.

However, for the DRSF approach, it seems that when multiple contextual variations were combined, the RR drastically dropped. For instance, when only the body reference frame was varied, the DSRF approach was still able to achieve high RRs up to 100\%. However, when the trials from the context \textit{Combined} had to be classified, the RR dropped drastically to below 50\%. Hence, when small variations in the body frame are the only source of contextual variation, approaches that are not right-invariant were still able to correctly classify the motions. However, when multiple contextual variations occurred simultaneously, the combined effect of multiple contextual dependencies led to severe confusion.
	
\section{Real-time Proof of Concept}
\label{sec:PoC}

This section gives a high-level overview of a real-time PoC for the online recognition of hand palm motion gestures for human-robot collaboration in ever-changing contexts. 

\begin{figure}[t]
	\centering
	\medskip
	\includegraphics[width=0.85\linewidth]{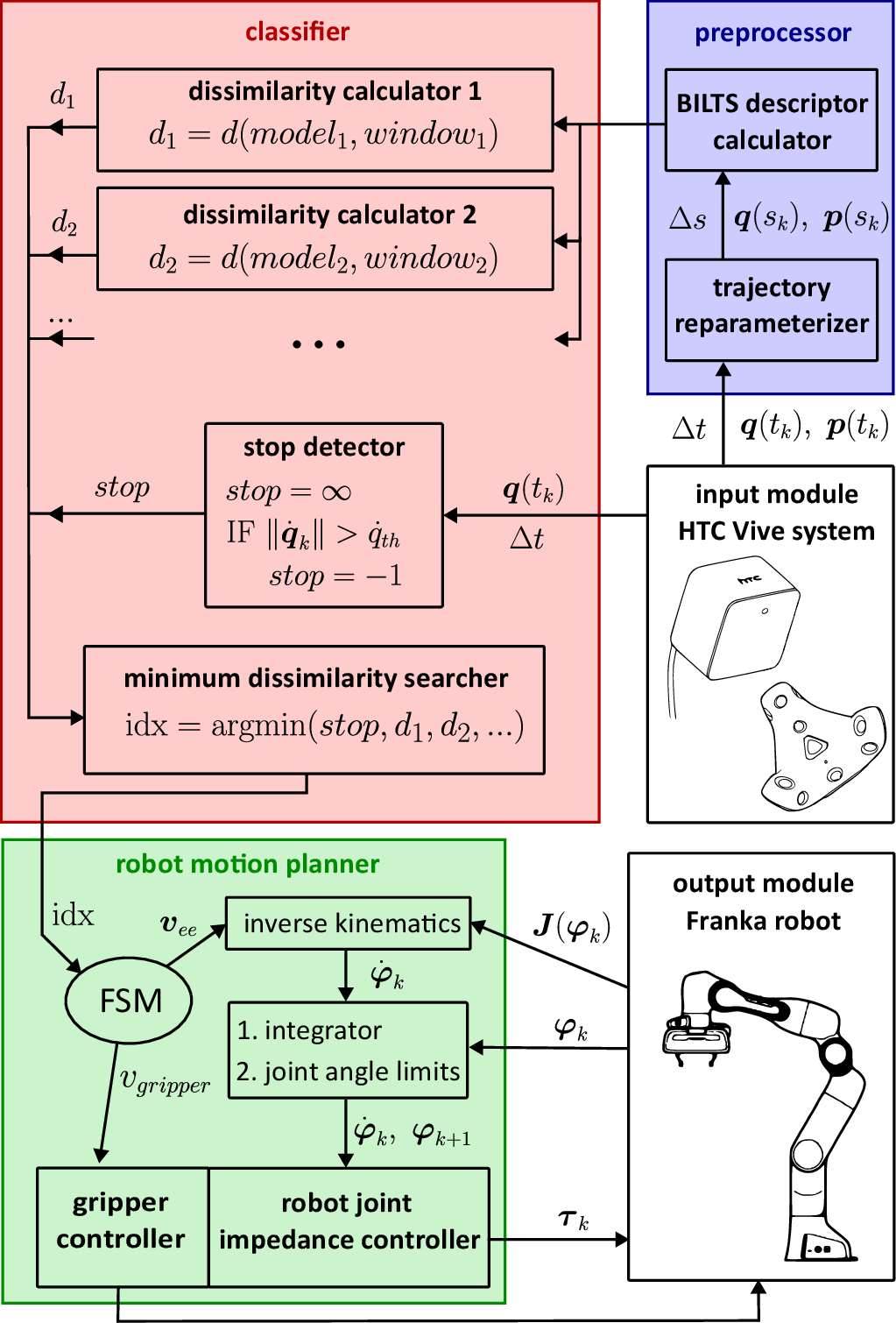}
	\caption{Flowchart of the developed real-time PoC. The flowchart starts with the HTC Vive system located in the middle right of the figure.}
	\label{fig:flowchart}
	\vspace{-5pt}
\end{figure}

Specifically, to robustly deal with contextual variations, the BILTS$^+$ invariant descriptor approach was leveraged and implemented, but now in an online fashion. The recognized gestures were mapped to motion plans for the Franka Panda robot. Figure~\ref{fig:flowchart} visualizes a flowchart of the PoC. The different components of this flowchart are detailed below.

\textit{Input module}: The input module was the HTC Vive system, consisting of a tracker and two base stations. The user's hand motion was captured by holding an HTC Vive tracker. The position and orientation of the tracker were represented as temporal position coordinates $\boldsymbol{p}(t_k)$ and quaternion coordinates $\boldsymbol{q}(t_k)$, with $t_k$ the time at sample $k$, for efficient data transmission. This format can always be converted to a homogeneous matrix representation $\boldsymbol{T}(t_k)$ and vice versa.

\textit{Preprocessor}: The real-time trajectory data was processed to BILTS$^+$ descriptor samples in an online fashion. This preprocessing included a trajectory reparameterizer, and a BILTS$^+$ descriptor calculator.

\textit{Trajectory reparameterizer}: 
To deal with time-profile variations, trajectory reparameterization was again implemented as a preprocessing step, but now in an online fashion, which is described on a high level in Algorithm~\ref{alg:reparameterization}. 

The reparameterizer resampled the temporal trajectory $\boldsymbol{T}(t_k)$ into a geometric trajectory $\boldsymbol{T}(s_i)$, with $s_i$ the geometric progress at sample $i$. The trajectory $\boldsymbol{T}(s_i)$ essentially represents a subsampled version of $\boldsymbol{T}(t_k)$. This subsampling ensured that the \textit{screw path} between successive samples, $\boldsymbol{T}(s_{i-1})$ and $\boldsymbol{T}(s_{i})$, always exceeded a given geometric progress resolution $\Delta s_{\text{res}}$. The screw path between samples was calculated using the formulas in \cite{verduyn2023enhancingmotiontrajectorysegmentation}.

\textit{BILTS$^+$ descriptor calculator}: After reparameterization, the BILTS$^+$ descriptor samples were computed in real time using the numerical calculation routine as detailed in~\cite{verduyn2025}.

\textit{Offline gesture modeling}:
The BILTS$^+$ descriptors of the prerecorded gestures from the context \textit{Original 1} of the HPM dataset were precomputed and stored as \textit{gesture models}.
Each model inherited the class label from its respective trial. 

To eliminate code duplication between offline and online preprocessing routines and to avoid inconsistencies between the low-level implementations of these routines, a single preprocessing routine (the online routine) was used for both offline gesture modeling and online recognition. Specifically, during offline gesture modeling, the prerecorded trials were simulated as if they were performed in real time, and the gesture models were calculated using the online routine.

\textit{Classifier}:
The gestures were recognized using a 1-NN classifier. This 1-NN classifier was implemented in a modular approach, consisting of multiple \textit{dissimilarity calculator} modules, one \textit{stop detector} module, and one \textit{minimum dissimilarity searcher} module. 

\textit{Dissimilarity calculator}: For each gesture model, a dedicated dissimilarity calculator module was defined, with the gesture model serving as the reference for dissimilarity measurement. In real time, the dissimilarity calculator module computed the average pointwise \textit{BILTS}$^+$ \textit{dissimilarity} $d$~\cite{verduyn2025} between the gesture model and a moving window of real-time BILTS$^+$ descriptor samples. Each module defined its own moving window,  with the window size matching the number of samples in the corresponding gesture model. 

To handle misalignments along the progress axis due to transient data in real-time (i.e., the gesture has started but has not yet finished), the cross-correlation between the model and the real-time data was computed to find the latency that resulted in the highest similarity (i.e., smallest dissimilarity) between the model and the real-time data. The considered latencies ranged from a single sample up to the nearest integer equivalent of 20\% of the total samples in the model. 

\begin{algorithm}[t]
	\caption{Online trajectory reparameterizer}
	\label{alg:reparameterization}
	\begin{algorithmic}[1]
		\STATE \textbf{Input:} Temporal trajectory $\boldsymbol{T}(t_k)$, threshold $\Delta s_{\text{res}}$
		\STATE \textbf{Output:} Geometric trajectory $\boldsymbol{T}(s_i)$
		\STATE \textbf{Initialize:} $ i \gets 0 $,  $ k \gets 0 $, $ \boldsymbol{T}(s_0) \gets \boldsymbol{T}(t_0) $
		\STATE \textbf{while}~receiving temporal trajectory samples \( \boldsymbol{T}(t_k) \)~\textbf{do}
		\STATE 
		~~~$\Delta s \gets calculate\_screw\_path \left(\boldsymbol{T}(s_i), \boldsymbol{T}(t_k)\right)$
		\STATE ~~~\textbf{if}~\( \Delta s > \Delta s_{\text{res}} \)~\textbf{then}
		\STATE ~~~~~~$i \gets i+1$ \\
		\STATE ~~~~~~$\boldsymbol{T}(s_i) \gets \boldsymbol{T}(t_k)$
	\end{algorithmic}
\end{algorithm}

An advantage of using \textit{local} descriptors when determining the progress latency between a model and a moving window of real-time data is that local descriptors allow computationally efficient updates as the window advances. Since previously computed local descriptor samples remain unchanged as the window advances, most of the previously computed local pointwise similarities with the model can be reused. Hence, an incremental algorithm can be devised and implemented. This is not the case for global descriptors, since all global descriptor samples have to be recomputed when the moving window advances. 

As shown in Figure~\ref{fig:flowchart}, each dissimilarity calculator model utilized a different moving window of real-time data, with window size equal to the number of samples of the corresponding gesture model. An alternative approach would be to use one single moving window of real-time BILTS$^+$ descriptor samples for all dissimilarity calculator modules, with the window size determined by some heuristic. DTW could then be used to address mismatches between the model and window lengths. However, this method might introduce bias, as it might favor gesture models whose lengths closely match the length of the moving window. 

\textit{Minimum dissimilarity searcher}: 
The dissimilarity calculators sent their BILTS dissimilarities $d$ to the minimum dissimilarity searcher, which then identified the lowest dissimilarity and retrieved the corresponding gesture class label. To avoid false positives, a recognition event was triggered only if this lowest dissimilarity fell below a predefined threshold. The tuning of this threshold involved balancing between false positive rejection and recall, where a higher threshold improves recall but also increases the risk of false positives. For the PoC, the threshold was manually tuned to be strict, hence favoring false positive rejection over recall.

\textit{Stop detector}:
A stop signal, triggered by a fast rotation of the hand, was also implemented. Fast rotations of the hand were detected by thresholding on the quaternion derivative. Upon detection, a stop signal of $d_{stop} = -1$ was sent to the minimum dissimilarity searcher. In this manner, the detection of a stop signal was treated similarly to the detection of any other gesture. Remark that all the dissimilarities from the dissimilarity calculator modules are positive by definition. Hence, when a stop signal is detected, $d_{stop} = -1$, this stop detection always took precedence.

\textit{Robot motion planner}:
The robot motion planner incorporated a Finite State Machine (FSM) which planned the end-effector velocity $\boldsymbol{v}_{ee}$ and gripper finger velocity $v_{gripper}$ based on the recognized gesture. In case of a detected stop motion, a graceful deceleration toward standstill was planned. Using the robot's Jacobian $\boldsymbol{J}$ at the current robot joint angles $\boldsymbol{\varphi}_k$, these end-effector velocity plans were mapped to joint velocities $\dot{\boldsymbol{\varphi}}_k$. These joint velocities where then integrated to obtain the joint positions $\boldsymbol{\varphi}_{k+1}$ for the next iteration, which were then passed to a low-level joint impedance controller, which finally computed the necessary joint torques $\tau_k$ for the Franka Panda robot (\textit{output module}).

\subsection*{Results of the implemented PoC}
\begin{figure}[t]
	\centering
	\medskip
	\includegraphics[width=0.75\linewidth]{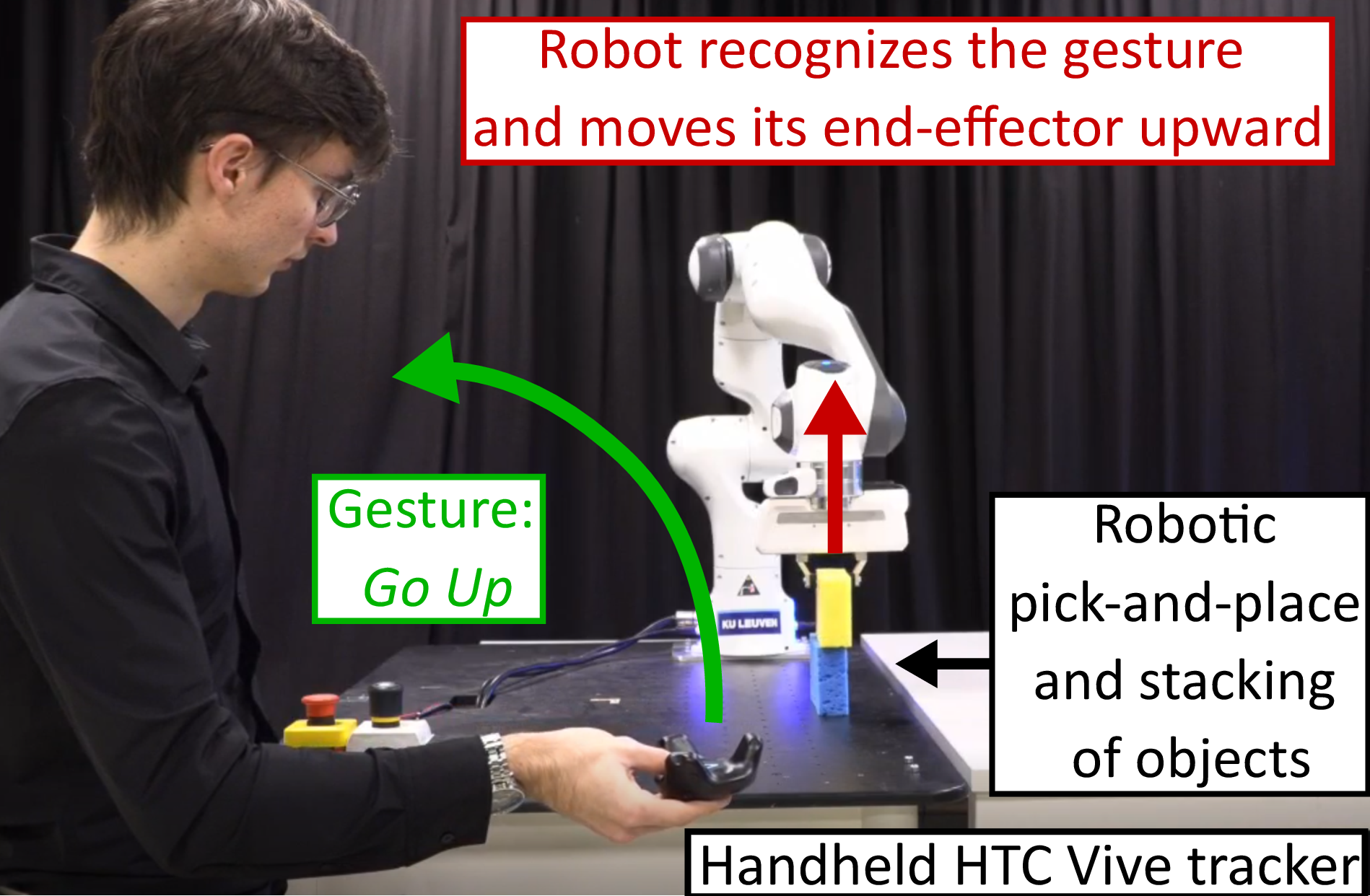}
	\caption{Video screenshot of the developed real-time PoC}
	\label{fig:screenshot}
	\vspace{0pt}
\end{figure}

The PoC was validated by using hand palm motion gestures to control a robotic pick-and-place and object-stacking task. A video showcasing this PoC is made accessible online\footnote{Video available on Zenodo : \\ \url{https://doi.org/10.5281/zenodo.14679070}.}. A screenshot of the video is shown in Figure~\ref{fig:screenshot}.

As is shown in the video, the user was free to move around (implicitly changing the world frame) and hold the HTC Vive tracker in arbitrary ways (implicitly changing the body frame) without compromising recognition accuracy. Hence, the implemented invariant descriptor approach (BILTS$^+$) effectively dealt with reference frame variations. 

Table~\ref{tab:results_realtime} lists the 21 gestures performed in the video, along with the timestamps of their initiation and whether each gesture was correctly detected. The robot failed to detect the \textit{Go Right} gestures at timestamps 1:32 and 2:31, likely due to insufficient similarity between the performed gestures and the gesture model. To proceed, the user performed the gesture again until it was correctly detected. 

The \textit{Go Down} gesture at timestamp 2:06 triggered the stop signal, likely because the gesture (involving a rotation of the hand) was performed too quickly. This hence demonstrates the system's safety feature, which defaults to \textit{stop} over gesture detection when faced with ambiguous inputs.

Hence, 21 gestures were performed, with 18 correctly detected. No false positives occurred, even though the user made arbitrary movements while moving around. This resulted in a high precision of $18/(18+0) = 100\%$, a recall of $18/(18+3) = 85.7\%$ , and a solid $F_1$-score of 92.3\%.

\section{Discussion and conclusion}
The primary goal of this work was to enhance the recognition of gestures (i.e. hand palm motions) by eliminating reference frame bias through the application of invariant descriptor approaches. To achieve this, a novel dataset of hand palm motion gestures was first presented. Then, from a set of state-of-the-art invariant descriptor approaches, the most suitable approach (i.e. BILTS$^+$) was identified. 
Finally, this approach was validated for the real-time recognition of hand palm motion gestures by developing an online PoC.

The approach does not rely on intensive training on large datasets. In contrast, for each gesture, only a few gestures recorded in one specific context were used. The reference frame dependency (or bias) was then removed during gesture modeling and recognition by leveraging the BILTS$^+$ invariant desciptor approach. This allowed the robust recognition of gestures in various and ever-changing contexts.
Furthermore, the frame-invariance of the approach eliminated the need for frame calibration: neither the calibration between the tracker frame and any anatomical frame of the hand, nor between the tracker frame and the sensor frame, nor between the sensor frame and the robot base frame was required. As a result, this frame-calibration-free approach significantly simplified setup deployment.

\definecolor{mygreen}{RGB}{0, 180, 0}
\begin{table}[t]
	\centering
	\caption{Timeline and detection results of the performed gestures.}
	\label{tab:results_realtime}
	\resizebox{0.9\linewidth}{!}{%
		\begin{tabular}{ccc|ccc}
			\textbf{time} & \textbf{gesture} & \textbf{correct} & \textbf{time} & \textbf{gesture} & \textbf{correct} \\ \midrule
			0:31 & \textit{Go Home} & \textcolor{mygreen}{\checkmark} & 2:06 & \textit{Go Down} & \textcolor{red}{\ding{55}} \\
			0:44 & \textit{Go Left} & \textcolor{mygreen}{\checkmark} & 2:11 & \textit{Go Down} & \textcolor{mygreen}{\checkmark} \\
			0:52 & \textit{Go Down} & \textcolor{mygreen}{\checkmark} & 2:19 & \textit{Close Gripper} & \textcolor{mygreen}{\checkmark} \\
			1:14 & \textit{Close Gripper} & \textcolor{mygreen}{\checkmark} & 2:26 & \textit{Go Up} & \textcolor{mygreen}{\checkmark} \\
			1:20 & \textit{Go Up} & \textcolor{mygreen}{\checkmark} & 2:31 & \textit{Go Right} & \textcolor{red}{\ding{55}} \\
			1:32 & \textit{Go Right} & \textcolor{red}{\ding{55}} & 2:33 & \textit{Go Right} & \textcolor{mygreen}{\checkmark} \\
			1:34 & \textit{Go Right} & \textcolor{mygreen}{\checkmark} & 2:42 & \textit{Go Down} & \textcolor{mygreen}{\checkmark} \\
			1:38 & \textit{Go Down} & \textcolor{mygreen}{\checkmark} & 2:48 & \textit{Open Gripper} & \textcolor{mygreen}{\checkmark} \\
			1:47 & \textit{Open Gripper} & \textcolor{mygreen}{\checkmark} & 2:58 & \textit{Go Up} & \textcolor{mygreen}{\checkmark} \\
			1:53 & \textit{Go Up} & \textcolor{mygreen}{\checkmark} & 3:05 & \textit{Go Home} & \textcolor{mygreen}{\checkmark} \\
			1:58 & \textit{Go Left} & \textcolor{mygreen}{\checkmark} & && \\
		\end{tabular}
	}\vspace{-10pt}
\end{table}

In the PoC, a motion to the left was signaled by a rectilinear translation of the hand, while a motion to the right was signaled by a circular motion of the hand. However, it might seem more intuitive to signal the directions explicitly by pointing left or right. However, this approach would require reference frame calibration to accurately interpret these directions. We emphasize that the low-level control of the robot using gestures was intended solely as a PoC.  The invariant approach is most suited for recognizing gestures commanding higher-level actions unrelated to specific reference frames or directions, such as the actions: \textit{start}, \textit{stop}, \textit{go slower}, \textit{go faster}, \textit{open gripper}, \textit{close gripper}, \textit{change controller}, \textit{repeat task}, \textit{switch task}, etc.

Similarly as in \cite{verduyn2025}, a \mbox{1-NN} classifier was used for the experimental comparative analysis of the different invariant descriptor approaches. The use of a basic 1-NN classifier allowed to reveal the root causes of limitations of the different approaches. Nevertheless, invariant descriptor approaches can readily be combined with more advanced classifiers from the literature. For instance, instead of learning models from raw trajectory coordinates, which are highly dependent on the choice of coordinate reference frames, models can be learned from invariant trajectory descriptors, where the descriptor calculation can be interpreted as a trajectory preprocessing step. Alternatively, learning from raw trajectory coordinates can be complemented by incorporating invariant descriptors or similarity measures as additional input features. 

\textbf{To conclude}, a novel publicly available dataset of hand palm motion gestures was introduced. Second, the most suitable invariant descriptor approach for this dataset was found to be BILTS$^+$. Third, a PoC of the BILTS$^+$ approach for online recognition under reference frame variations was developed and validated. %
This work demonstrated the effectiveness of the invariant descriptor approach as a standalone solution. Moreover, we believe our methods can also be utilized within other state-of-the-art recognition systems to improve their robustness against contextual variations.

\section*{acknowledgments}
This work was supported by a project that has received funding from the European Research Council (ERC) under the European Union's Horizon 2020 research and innovation programme (Grant agreement No. 788298). 

ChatGPT was used to assist in improving the grammar, style, and tone of the text.

\bibliographystyle{IEEEtran}
\bibliography{allpapers}

\end{document}